\documentclass{article}

\usepackage{microtype}
\usepackage{graphicx}
\usepackage{booktabs} 
\usepackage{subfig}
\usepackage{caption}
\usepackage{hyperref}

\usepackage[accepted]{icml2024}

\usepackage{amsmath}
\usepackage{amssymb}
\usepackage{mathtools}
\usepackage{amsthm}
\usepackage[vlined, linesnumbered, algo2e]{algorithm2e}
\usepackage{lipsum}

\theoremstyle{plain}

\theoremstyle{definition}

\theoremstyle{remark}

\usepackage[textsize=tiny]{todonotes}

\icmltitlerunning{DocParseNet: Advanced Semantic Segmentation and OCR Embeddings for
Efficient Scanned Document Annotation}

\providecommand{\keywords}[1]{\textbf{\textit{Keywords: }} #1}

\begin{document}

\twocolumn[
\icmltitle{DocParseNet: Advanced Semantic Segmentation and OCR Embeddings for Efficient Scanned Document Annotation}




\begin{icmlauthorlist}
\icmlauthor{Ahmad Mohammadshirazi}{yyy}
\icmlauthor{Ali Nosrati Firoozsalari}{yyy}
\icmlauthor{Mengxi Zhou}{yyy}
\icmlauthor{Dheeraj Kulshrestha}{sch}
\icmlauthor{Rajiv Ramnath}{yyy}

\end{icmlauthorlist}

\icmlaffiliation{yyy}{Department of Computer Science and Engineering, The Ohio State University, Columbus, OH 43210, USA}
\icmlaffiliation{sch}{Flairsoft Company, Columbus, OH 43235, USA}

\icmlcorrespondingauthor{Ahmad Mohammadshirazi}{mohammadshirazi.2@osu.edu}

\icmlkeywords{Machine Learning, ICML}

\vskip 0.3in
]
\makeatletter\def\Hy@Warning#1{}\makeatother


\printAffiliationsAndNotice{} 


\begin{figure*}[ht]
  \centering
  \includegraphics[width=1\linewidth]{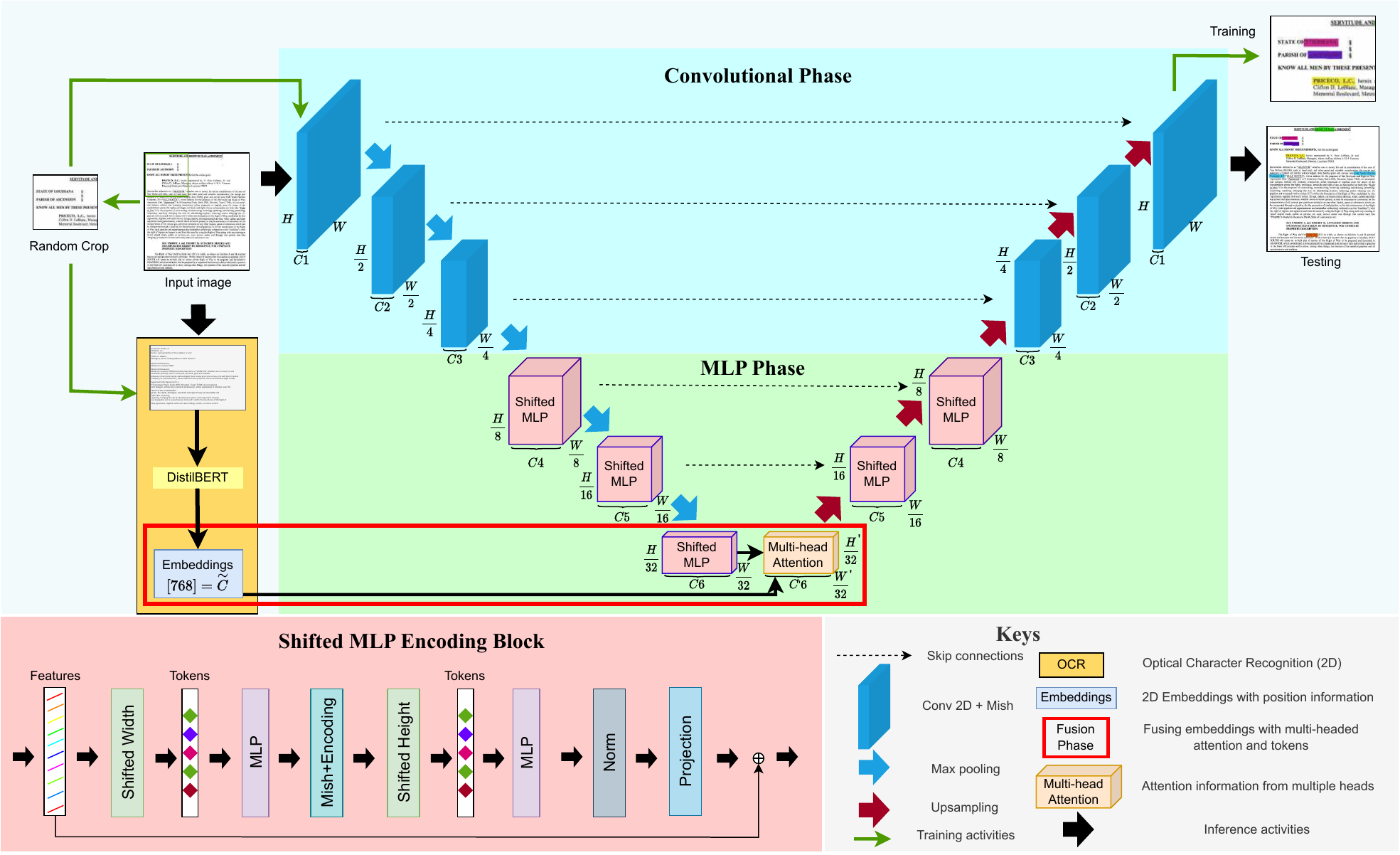}
  
  \caption{Diagram illustrating the DocParseNet architecture, emphasizing its four core modules: Pixel Level Module, utilizing convolutional layers for detailed visual feature extraction; MLP phase, distilling broader image features; Text Feature Module, employing OCR for textual data extraction and contextual understanding; and Fusion Module, synergistically integrating multi-modal features to enhance document parsing and prediction accuracy.}
  \label{fig:proposed}
\end{figure*}

\begin{abstract}
Automating the annotation of scanned documents is challenging, requiring a balance between computational efficiency and accuracy. DocParseNet addresses this by combining deep learning and multi-modal learning to process both text and visual data. This model goes beyond traditional OCR and semantic segmentation, capturing the interplay between text and images to preserve contextual nuances in complex document structures. Our evaluations show that DocParseNet significantly outperforms conventional models, achieving mIoU scores of 49.12 on validation and 49.78 on the test set. This reflects a 58\% accuracy improvement over state-of-the-art baseline models and an 18\% gain compared to the UNext baseline. Remarkably, DocParseNet achieves these results with only 2.8 million parameters, reducing model size by approximately 25 times and speeding up training by 5 times compared to other models. These metrics, coupled with a computational efficiency of 0.039 TFLOPs (BS=1), highlight DocParseNet's high performance in document annotation. The model's adaptability and scalability make it well-suited for real-world corporate document processing applications. The code is available at https://github.com/ahmad-shirazi/DocParseNet.
\end{abstract}


\graphicspath{ {../} }
\section{Introduction}
\label{sec:intro}

Achieving efficient and precise annotation of data from scanned documents remains a formidable challenge in machine learning and computer vision.
\cite{dutta2024visformers,o2020deep}. These scanned images, rich with textual and graphical content, are crucial across various fields.
Despite advancements in Optical Character Recognition (OCR) \cite{li2023trocr} and image processing, a significant gap persists between document digitization and true semantic understanding \cite{borchmann2021due}.

This gap is further widened by the complex nature of documents, where the structure of characters and their contextual meaning are intricately intertwined. Conventional methods like document layout analysis offer structural parsing \cite{shen2021layoutparser, zhou2022structure}, while semantic analysis aims to uncover deeper meanings \cite{mercha2023machine}. Although template-based extraction efficiently harvests data from predefined fields in structured documents, these methods are insufficient for documents blending structured and unstructured data. The variability in document quality, format diversity, and layout complexity further challenges current methods, especially with documents exhibiting high variance in text density, font styles, and graphical elements, as well as poor scanning quality \cite{zhou2023transformer}. Additionally, the computational demand and need for large, annotated datasets for training models exacerbate these challenges, creating a bottleneck for efficient data extraction and processing \cite{lepcha2023image, liang2022advances}.

A critical observation in current methodologies is their tendency to rely predominantly on either text or image data, often overlooking the comprehensive potential of utilizing both \cite{saharia2022photorealistic}. Solely text-based approaches can fail to appreciate the intricate layout and structural nuances of documents, while image-based methods might inadequately capture textual details. 
Multi-modal learning, which combines diverse data types such as images and text, significantly enhances the precision of automated annotation for complex tasks, albeit with higher computational costs.
\cite{liu2024textmonkey,huang2022layoutlmv3,xu2021layoutxlm}.


Recognizing this challenge, we introduce DocParseNet, a model specifically designed to integrate textual and visual data, leveraging the strengths of each modality to address the limitations inherent in single-type information processing. DocParseNet's architecture synergizes semantic segmentation with OCR embeddings to achieve comprehensive document annotation, not only increasing accuracy but also reducing computational costs.

The remainder of this paper is structured as follows: Section 2 reviews related work, Section 3 outlines the DocParseNet methodology, Section 4 presents experimental results and discussion, and Section 5 concludes with future work.


\section{Related Work}

Semantic segmentation, a cornerstone task in computer vision, necessitates a nuanced understanding of image details at multiple scales. Notable semantic segmentation models like SegFormer \cite{xie2021segformer}, LawinTransformer \cite{yan2022lawin}, and MaskFormer \cite{cheng2021per}, utilizing multi-scale feature extraction critical for intricate scene segmentation.

The UNet framework employs an encoder-decoder architecture with skip connections to preserve spatial information \cite{huang2020unet, navard2024probabilistic}. This approach benefits from convolutional neural networks' (CNNs') local feature extraction capabilities. Inspired by the UNet framework, UNeXt \cite{valanarasu2022unext} offers a less computationally intensive model by utilizing a Multi-Layer Perceptron (MLP) approach \cite{taud2018multilayer}.

Building on the concept of global and local feature extraction exemplified by models like MedT \cite{valanarasu2021medical}, VT-Unet \cite{peiris2022robust}, and TransBTS \cite{wang2021transbts}, our model employs an efficient random window crop mechanism \cite{zhong2020random}.
The integration of an innovative OCR stage for textual embeddings, combined with shifted MLPs and multi-head attention, enhances semantic comprehension \cite{fang2024single, huang2022layoutlmv3}.

Multi-modal learning, which merges vision and language, represents a vital research direction \cite{uppal2022multimodal}. In document analysis, this approach enables systems to assimilate and interpret visual content alongside text, achieving a unified comprehension that surpasses uni-modal methods \cite{zhang2024m2doc}.

Significant advancements include models like CLIP \cite{zhu2023pointclip}, which uses a Vision Transformer (ViT) \cite{chen2023cf} for visual model learning from natural language supervision, and BLIP, which employs a ViT encoder for both understanding and generation tasks within vision-language pre-training \cite{li2023blip}. In document understanding, models like LayoutLM \cite{huang2022layoutlmv3} and Docformer \cite{appalaraju2024docformerv2} integrate text, layout, and image information to achieve state-of-the-art performance on visually-rich document tasks.

DocParseNet builds on these advancements by employing a modified UNet for visual feature extraction and DistilBERT \cite{sanh2019distilbert} for text embedding, ensuring efficiency without sacrificing language understanding. A specialized fusion module integrates these modalities seamlessly, providing an advanced multi-modal learning system with improved document interpretation capabilities.

\begin{table*}[t]
    \caption{Test Set Performance and Efficiency of DocParseNet}
    \centering
        \begin{tabular*}{0.95\textwidth}{@{\extracolsep{\fill}}  l | r r r r r r | c}
            \hline
            \textbf{Method} & \textbf{mIoU} & \textbf{AT} & \textbf{State} & \textbf{County} & \textbf{Grantor} & \textbf{Grantee} &  \textbf{TFLOPs}\\
            \hline
                     
             Segformer-B5 \cite{xie2021segformer} & 34.81 & 37.36 & 38.72 & 44.28 & 8.81 & 44.87 &  0.39 \\ 
             
             UNext \cite{valanarasu2022unext} & 42.04 & 36.16 & 53.81 & 52.01 & 24.48 & 43.73 &  0.06 \\
            \hline
            
             DocParseNet & \textbf{49.78} & \textbf{43.06} & \textbf{65.66} & \textbf{53.90} & \textbf{36.14} & \textbf{50.12} &  \textbf{0.04}  \\
            \hline
        \end{tabular*}
       
    \label{table1}
    
    \footnotesize{$^*$ AT: Agreement Title. For each field, we mark the highest score in bold.}
\end{table*}


\section{DocParseNet}
Our proposed model, DocParseNet (Figure \ref{fig:proposed}), is a multi-modal deep learning architecture designed for parsing and annotating corporate agreement PDFs. It processes images derived from these PDFs, identifying key elements such as State, County, Agreement Title, Grantee Company, and Grantor Company. We trained DocParseNet on a dataset of 4,102 expert-annotated corporate agreement PDFs, comprising 19,821 images, split into training, validation, and testing sets using an 8-1-1 ratio. The non-static positioning of fields, influenced by the dynamic document structure, adds further complexity to the model's task. This heterogeneity, however, is essential, enriching the dataset and providing a robust foundation for efficient DocParseNet training.





\subsection{DocParseNet\_UNet}

DocParseNet\_UNet comprises a modified Convolutional Neural Network (CNN) for image processing and a Shifted MLP Encoding block for enhancing image features. This architecture is depicted in Figure \ref{fig:proposed}.


\textit{CNN\-Conv 2D+Mish:}
The CNN component of DocParseNet\_UNet as demonstrated by equation \ref{eq.CNN} uses a series of 2D convolutional layers to extract hierarchical spatial features from input images, utilizing a 1x1 convolution (conv1x1) to refine feature maps between stages. The CNN employs the Mish activation function \cite{misra2019mish}, which preserves more information compared to traditional activation functions like ReLU. Equation \ref{eq.rand} takes the input image (I) and outputs a random crop (I$'$). The random crop is then passed onto the aforementioned CNN component, and the output is calculated.


\begin{equation}
\label{eq.CNN}
    X_{\text{shift}} = f_{\text{CNN}}(I')
\end{equation}
\begin{equation}
\label{eq.rand}
    I' = f_{\text{Random\_crop}}(I)
\end{equation}



\textit{Shifted MLP Encoding Block:}
The Shifted MLP Encoding Block is designed to overcome the limitations of CNNs, particularly their localized receptive fields. By employing a shift operation in the MLP’s receptive window across the feature map, it achieves cross-window connectivity, capturing broader contextual information efficiently. This structure enriches feature representation with contextual information spanning both local and distant regions of the input. The shifted MLP can be described as the following:


\begin{equation}
\label{eq.shift}
    V_n = \text{Linear}\left( \text{Mish}\left( \text{DWConv}\left( W_1 \cdot X_{\text{shift}} + b_1 \right) \right) \right),
\end{equation}


where a Mish activation function is applied after Depth-Wise Convolution (DWConv) and a feed-forward neural network. Another feed-forward neural network is then applied to the output. Additional details are provided in Appendix \ref{appendix1-mlp}.



\subsection{DocParseNet\_OCR}


The OCR phase in DocParseNet, shown in Figure \ref{fig:proposed}, is designed for text recognition within document images. It uses optical character recognition to convert visual text into machine-encoded text, enabling the extraction of textual embeddings. This study utilizes a pre-trained DistilBERT model for its efficient text encoding capabilities. The OCR process includes Image-to-Text Conversion ($f_{\text{OCR}}$), text embedding using the Distilbert model ($f_{\text{BERT}}$), and [CLS] Token Embedding (E). For further details, please refer to Appendix \ref{appendix1-ocr}.


\begin{equation}
    E = f_{\text{BERT}}(f_{\text{OCR}}(I'))
\end{equation}



\subsection{DocParseNet\_Fusion}

The fusion phase in DocParseNet (Figure \ref{fig:proposed}) combines visual and textual information using multi-head attention. This mechanism aligns and merges features from the UNet and OCR streams, leveraging their complementary strengths for accurate document annotation.

In DocParseNet, the fusion process combines the last shifted MLP encoding block output with BERT embeddings via multi-head attention:

- Let \( V_6 \in \mathbb{R}^{B \times C_6 \times \frac{H}{32} \times \frac{W}{32}} \) be the output from the last shifted MLP encoding block.

- Let \( E \in \mathbb{R}^{B \times 768} \) be the embedding from the DistilBERT model.

The multi-head attention outputs a feature map \( F \in \mathbb{R}^{B \times C'_6 \times \frac{H'}{32} \times \frac{W'}{32}} \) by:


\begin{equation}
F = f_{\text{MultiHead}}(V_6, E).
\end{equation}


Placing the fusion at the UNet bottleneck integrates deep visual features with rich textual embeddings, creating a dense feature representation crucial for precise annotation, especially with limited training data.

DocParseNet combines UNet-based visual segmentation with OCR-based text recognition, enhanced by a fusion module for visual and textual modalities, to achieve high annotation precision. This process is detailed in Appendix \ref{appendix1-algorithm}, Algorithm \ref{alg:docparsenet_algorithm}, demonstrating the model's robustness and efficiency (Figure \ref{fig:proposed}). The hardware setup for experimentation is explained in Appendix \ref{appendix1-harware}.



\section{Result}


\begin{figure}[ht]
    \centering
    \includegraphics[width=0.48\textwidth]{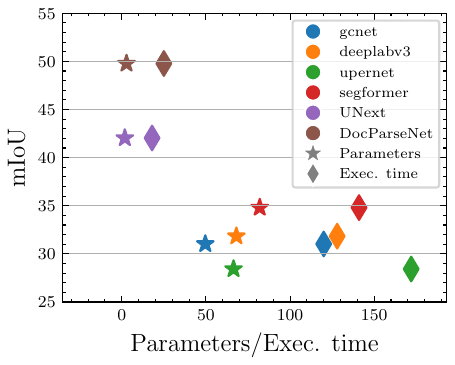}
    \caption{Parameters and Execution time}
    \label{fig:fig-3-pra-exec}
\end{figure}


In our empirical study, we evaluated the proposed DocParseNet model by benchmarking it against leading semantic segmentation models, including GCNet \cite{cao2020global}, UPerNet \cite{xiao2018unified}, DeeplabV3 \cite{chen2017rethinking}, and Segformer \cite{xie2021segformer}, using the segmentation \cite{mmseg2020, perera2024segformer3d} framework. These state-of-the-art models, optimized for image feature extraction, do not inherently account for text-based information.

The comparative analysis, detailed in Table \ref{table1} and Appendix \ref{appendix2-IoUs} (Table \ref{table2-appendix} and Figure \ref{fig:ious}), highlights DocParseNet's enhanced ability to incorporate text information, resulting in a significant performance boost. Our model's superior performance is evidenced by its mean Intersection over Union (mIoU) scores of 49.12 on the validation set and 49.78 on the test set. This demonstrates its robustness and superior generalization in handling the challenges of semantic segmentation and OCR embedding for scanned document annotation.
Additionally, the training dynamics of DocParseNet, as shown in  Appendix \ref{appendix2-Training}, Figure \ref{fig:train_validation}, exhibit robust learning and generalization capabilities through a consistent upward trajectory in IoU scores and a decrease in loss, highlighting its resistance to overfitting.

 As shown in Table \ref{table1}, Table \ref{table2-appendix}, and Figure \ref{fig:ious} in Appendix \ref{appendix2-Training}, DocParseNet achieved IoU scores of 39.77 on validation and 36.14 on testing for the 'Grantor' category, demonstrating a smaller discrepancy between validation and testing compared to baseline models. This robustness is due to our model's adept handling of varying boundary box locations across documents. The integration of OCR embeddings allows DocParseNet to maintain high accuracy despite spatial inconsistencies, showcasing its robustness and generalization capabilities. In the 'State' and 'County' categories, DocParseNet achieved exceptional performance with scores of 65.66 and 53.90, respectively. In the 'Agreement Title (AT)' and 'Grantee' categories, it maintained robust accuracy with scores of 43.06 and 50.12, respectively.


This adaptability is further enhanced by an empirically optimized channel configuration. Through extensive experimentation using Bayesian optimization, we determined the optimal channel sizes at each stage of the model to be [16, 32, 64, 128, 256, 320]. This configuration effectively captures the complex hierarchies of features necessary for accurate document segmentation.

Figure \ref{fig:fig-3-pra-exec} and Table \ref{table3-appendix} in Appendix \ref{appendix2-Computational} illustrate the performance evaluation of our baseline models and the proposed DocParseNet model. We utilize the HARP profiling framework \cite{vallabhajosyula2022towards} to gather model performance statistics, including Floating Point Operations (FLOPs) for a training step. A thorough examination consistently demonstrates DocParseNet's superior performance in accuracy and execution efficiency compared to all baselines.
DocParseNet achieves this by efficiently leveraging reduced parameters, leading to GPU memory savings, which allows for increased batch sizes or model fusion without compromising memory requirements or training latency. This efficiency enables our model to attain superior accuracies (Table \ref{table1}) with reduced training latency, without incurring additional memory costs compared to competitive baselines.


\section{Conclusion and Future}
In summary, our work with DocParseNet establishes a new standard in document annotation, demonstrating exceptional performance and efficiency in semantic segmentation and OCR embedding for scanned documents. Our innovative framework, which utilizes deep learning and multi-modal learning, outperforms existing methods, evidenced by impressive mIoU scores. The accelerated training enhances reusability and facilitates frequent updates within a feedback loop, significantly reducing manual annotation efforts and improving accuracy. DocParseNet's economical computational resource usage and enhanced training speed highlight its potential for broad practical applications, particularly in domains requiring precise document analysis.
Moving forward, we aim to expand the scope of DocParseNet to encompass additional domains in document analysis, leveraging the latest advancements in deep learning technology. Our focus will be on scaling up pre-trained models, enabling them to utilize extensive training data and further advancing SOTA results to enhance the applicability of our models in real-world business scenarios within the Document AI industry.

\section*{Acknowledgements}
We extend our sincere gratitude to FlairSoft Company for their support. Additionally, we acknowledge the contributions of the OSU CETI Lab, with special thanks to Anuja Dixit and Manikya Swathi Vallabhajosyula, for their invaluable assistance.

\nocite{langley00}

\bibliography{docparsenet}
\bibliographystyle{icml2024}

\newpage
\appendix
\onecolumn
\section{More Info-DocParseNet}
\label{appendix1}

\subsection{Shifted MLP Encoding Block}
\label{appendix1-mlp}
Given an input feature map \( x \in \mathbb{R}^{B \times C \times H \times W} \), where \( B \) is the batch size, \( C \) is the channel count, and \( H, W \) are the feature map's height and width, the Shifted MLP Encoding block executes the following operations:

\begin{enumerate}
    \item A shift operation is applied along spatial dimensions (width) to rearrange the input tensor elements:
    \[ X_{\text{shift}} = \text{Shift}(x) \]
    where \( \text{Shift}(\cdot) \) cyclically shifts elements within each channel.
    
    \item The shifted output is processed through a linear layer:
    \[ x' = (W_1 \cdot X_{\text{shift}} + b_1) \]
    where \( W_1 \in \mathbb{R}^{C \times C'} \), \( b_1 \in \mathbb{R}^{C'} \) are the weights and biases of the first linear layer, and \( C' \) is the hidden feature dimension.
    
    \item Integration of Depth-Wise Convolution (DWConv) and Mish Activation Function is applied for local spatial interaction:
    \[  x'' = \text{Mish}(\text{DWConv}(x') )\]

    
    \item Dropout is then applied to the output of the Mish activation.
    \item A shift operation is applied along spatial dimensions (height).
    \item Finally, a second linear transformation is applied:
    \[ x''' = W_2 \cdot x'' + b_2 \]
    where \( W_2 \in \mathbb{R}^{C' \times C} \) and \( b_2 \in \mathbb{R}^{C} \) are the weights and biases of the second linear layer.
\end{enumerate}

\subsection{OCR Component}
\label{appendix1-ocr}
The OCR component of DocParseNet converts the textual content within document images into a format suitable for embedding by a language model. This process includes the following steps:

\begin{enumerate}
    \item The image-to-text conversion, typically performed by an OCR system such as Tesseract, can be denoted as a function \( \text{OCR}(\cdot) \) that maps an image \( I \) to a string of text \( T \):
    \begin{equation}
    T = \text{OCR}(I)
    \end{equation}

    \item The extracted text \( T \) is encoded into embeddings by a pre-trained language model such as DistilBERT. Let the DistilBERT model be represented as \( \mathcal{B}(\cdot) \). The process of obtaining the textual embeddings \( E \) is:
    \begin{equation}
    E = \mathcal{B}(T)
    \end{equation}
    which results in \( E \in \mathbb{R}^{B \times 768} \), where \( B \) is the batch size.

    \item The [CLS] token embedding, which serves as the aggregate representation of the input text, is extracted from the DistilBERT output to be used for downstream tasks:
    \begin{equation}
    E_{\text{[CLS]}} = E[:, 0, :]
    \end{equation}
    assuming that \( E \) is ordered such that the [CLS] token is the first in the sequence of embeddings.
\end{enumerate}

\subsection{DocParseNet Algorithm}
\label{appendix1-algorithm}

To encapsulate, DocParseNet employs a cohesive methodology as delineated in Algorithm \ref{alg:docparsenet_algorithm}, which synergizes UNet-based visual feature segmentation with OCR-based text recognition. This integration is further enhanced by a fusion module adept at amalgamating the visual and textual modalities, yielding a harmonized system with heightened annotation precision. Such a synergistic approach is fundamental to DocParseNet’s capability to deliver exceptional performance metrics, particularly notable given the limited training dataset. Algorithm \ref{alg:docparsenet_algorithm} outlines the strategic process flow that enables DocParseNet to maintain high accuracy in document image analysis, demonstrating the model's robustness and efficacy, with its components and processes visualized in Figure \ref{fig:proposed}.
\begin{algorithm}
\SetAlgoLined
\caption{DocParseNet Algorithm}
\label{alg:docparsenet_algorithm}

\KwIn{Document image dataset}
\KwOut{Set of annotated images with labels}

Initialize DocParseNet with UNet and OCR components\;

\For{Each document image $I$ in the dataset}{
    // UNet-based Visual Feature Extraction\\
    $I' \leftarrow \text{Random crop}(I)$\\
    $V \leftarrow \text{CNN\_Processing}(I')$\\
    $V \leftarrow \text{Apply Shifted MLP Encoding to } V$\\
    
    // OCR Text Extraction\\
    $T \leftarrow \text{OCR}(I')$\\
    $E \leftarrow \text{DistilBERT\_Encoding}(T)$\\
    
    // Multi-Modal Fusion\\
    $F \leftarrow \text{MultiHeadAttention}(V, E)$\\
    $R \leftarrow \text{LayerNormalization}(F + V)$\\
    
    Output annotated image with labels from $R$\;
}
\KwRet{Set of annotated images}
\end{algorithm}
\subsection{Hardware Setup for Experimentation}
\label{appendix1-harware}
We conduct experiments on the Ascend cluster\footnote{\url{https://www.osc.edu/resources/technical_support/supercomputers/ascend}} at the Ohio Supercomputer Center (OSC) using single-node, single-GPU setups. Each node is equipped with 12 CPU cores, featuring AMD EPYC 7643 (Milan) processors clocked at 2.3 GHz, with a maximum memory utilization of 10GB per core (totaling 120GB across the cores). The GPU employed is the NVIDIA A100 with 80GB memory. Please refer to the Results section for insights into the performance of our model.


\section{Additional Results}
\label{appendix2}

\subsection{Training Dynamics}
\label{appendix2-Training}

\begin{figure}[H]
\renewcommand\thefigure{B1}
    \centering
    \subfloat[IoU scores during training and validation phases.]{%
        \includegraphics[width=0.47\textwidth]{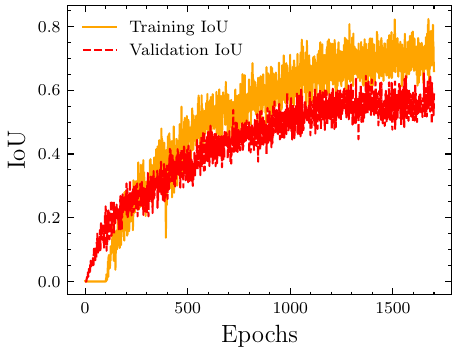}
        \label{fig:train_validation1}
    }
    \hfill
    \subfloat[Training and validation loss curves over epochs.]{%
        \includegraphics[width=0.47\textwidth]{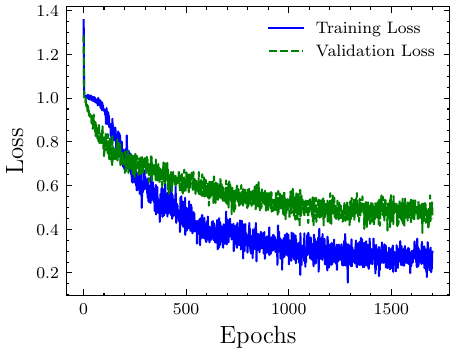}
        \label{fig:train_validation2}
    }
    \caption{Performance metrics of DocParseNet across 1700 epochs in the 'State' category. Subfigure (a) presents the IoU scores, while (b) illustrates the loss curves.}
    \label{fig:train_validation}
\end{figure}

The training dynamics of DocParseNet, as shown in Figure \ref{fig:train_validation}, are characterized by a consistent upward trajectory in IoU scores for the 'State' category, affirming the model's capacity to learn and refine its segmentation capabilities over time. The convergence of training and validation IoU, alongside a corresponding decrease in loss, illustrates the model's ability to generalize beyond the training data. This performance is underscored by the minimal gap between training and validation metrics, highlighting the robustness of DocParseNet against overfitting and its adeptness in handling the variegated nature of document structures within the 'State' category.

\begin{table*}[ht]
\renewcommand\thetable{B1}
    \caption{Performance (IoU scores) of the proposed DocParseNet method against other competitive methods on the [training] / validation / (test set).}
    \centering

        \begin{tabular*}{0.95\textwidth}{@{\extracolsep{\fill}}  l | r r r r r r }
            \hline
            \textbf{Method} & \textbf{mIoU} & \textbf{AT} & \textbf{State} & \textbf{County} & \textbf{Grantor} & \textbf{Grantee}\\
            \hline
             & [55.20] & [59.13] & [46.88] & [47.75] & [63.48] & [58.77] \\
             GCNet \cite{cao2020global} & 30.18 & 26.29 & 32.72 & 34.63 & 22.47 & 34.78 \\
             & (31.03) & (28.21) & (36.61) & (44.62) & (11.11) & (34.62) \\ [0.5em]
             
             & [47.29] & [53.76] & [37.70] & [36.99] & [58.02] & [49.97] \\
            UPerNet \cite{xiao2018unified} & 25.98 & 26.22 & 25.65 & 25.94 & 19.28 & 32.83 \\             
             & (28.43) & (28.98) & (28.49) & (40.80) & (11.01) & (32.87) \\ [0.5em] 
             
             & [54.65] & [56.70] & [45.91] & [48.37] & [63.77] & [58.48] \\
             DeeplabV3 \cite{chen2017rethinking} & 30.16 & 28.36 & 31.36 & 30.82 & 23.98 & 36.26 \\
             & (31.84) & (29.08) & (35.62) & (44.40) & (12.06) & (38.06) \\[0.5em] 
             
             & [68.26] & [69.68] & [60.63] & [63.90] & [75.75] & [71.33] \\
             Segformer-B5 \cite{xie2021segformer} & 35.23 & 34.53 & 34.78 & 39.69 & 30.67 & 36.48 \\
             & (34.81) & (37.36) & (38.72) & (44.28) & (8.81) & (44.87) \\ [0.5em]
             
             & [70.16] & [65.41] & [71.09] & [69.83] & [73.96] & [70.49] \\
             UNext \cite{valanarasu2022unext} & 41.65 & 35.53 & 51.47 & 47.69 & 31.51 & 42.04 \\
             & (42.04) & (36.16) & (53.81) & (52.01) & (24.48) & (43.73) \\
            \hline
             & [71.79] & [69.89] & [78.13] & [71.27] & [68.06] & [71.59] \\
             DocParseNet & \textbf{49.12} & \textbf{42.23} & \textbf{60.32} & \textbf{52.68} & \textbf{39.77} & \textbf{50.58} \\
             & \textbf{(49.78)} & \textbf{(43.06)} & \textbf{(65.66)} & \textbf{(53.90)} & \textbf{(36.14)} & \textbf{(50.12)} \\
            \hline
        \end{tabular*}
       
    \label{table2-appendix}
    
    \footnotesize{$^*$ AT: Agreement Title. For each field, we mark the highest score in bold.}
\end{table*}
\begin{table*}[ht]
\renewcommand\thetable{B2}
\centering
\caption{Performance metrics for our experiments in terms of memory usage and per-epoch execution time.}
\label{tab:my-table}
\begin{tabular}{lrrrrr}
\hline
\multicolumn{1}{l|}{\textbf{Model}} & \textbf{Params (M)} & \textbf{CPU Mem (GB)} & \multicolumn{1}{c}{\textbf{GPU Mem (GB)}} & \textbf{Model TFLOPs} & \textbf{Per Epoch (secs)} \\
\multicolumn{1}{c|}{} &  & max & max & (BS=1) & avg. \\ \hline
\multicolumn{1}{l|}{GCNet} & 49.62 & 2.61 & 12.20 & 1.1867 & 120 \\
\multicolumn{1}{l|}{UPerNet} & 66.41 & 2.62 & 13.20 & 1.4207 & 172 \\
\multicolumn{1}{l|}{DeeplabV3} & 68.1 & 2.62 & 12.60 & 1.6184 & 128 \\
\multicolumn{1}{l|}{Segformer-B5*} & 81.97 & 2.85 & 11.5 & 0.3867 & 141 \\
\multicolumn{1}{l|}{UNext} & 1.8 & 1.62 & 8.56 & 0.0546 & 18 \\ \hline
\multicolumn{1}{l|}{DocParseNet} & 2.8 & 6.06 & 9.30 & 0.0391 & 25 \\ \hline
\end{tabular}
\label{table3-appendix}
\footnotesize{ \break All experiments utilize a batch size of 8, except for '*' (with a batch size of 4), and the maximum memory reflects this specific batch size requirement. TFLOPs are calculated for a batch size of 1.}
\end{table*}

\pagebreak
\subsection{Comparative Analysis of DocParseNet's Superior IoU Performance}
\label{appendix2-IoUs}

\begin{figure*}
\renewcommand\thefigure{B2}
  \centering
  \includegraphics[width=16.5cm, height=9.5cm]{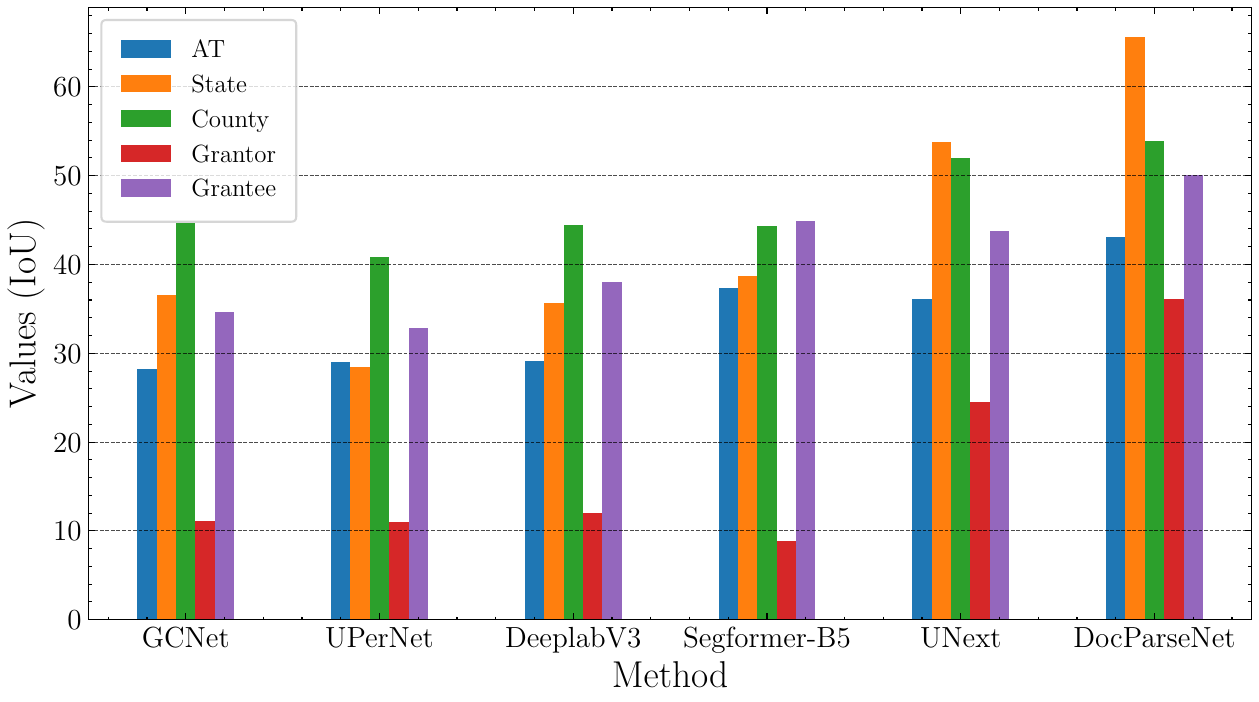}
  \caption{Assessment of DocParseNet compared to baseline methods across multiple datasets using IoU.}
  \label{fig:ious}
\end{figure*}

We provide a detailed comparative analysis of IoU scores, as shown in Table \ref{table2-appendix} and Figure \ref{fig:ious}, to evaluate the performance of DocParseNet relative to other advanced models. The results demonstrate that DocParseNet consistently outperforms the benchmarks set by GCNet, UPerNet, DeeplabV3, Segformer-B5, and UNext, with notable mIoU scores of 49.12 in validation and 49.78 in testing. Particularly in the 'State' and 'County' training categories, DocParseNet excels with scores of 65.66 and 53.90, respectively, showcasing its capability to effectively interpret the complex features of RoW documents. In the challenging 'Grantor' and 'Grantee' categories, DocParseNet maintains robust accuracy, with scores of 39.77 and 50.58, illustrating its proficiency in handling text location variability and document structural intricacies. The highest scores achieved by DocParseNet, highlighted in bold within the table, affirm its superior performance in extracting document information, substantiating the model's efficacy as detailed in the main body of this paper.

\subsection{Computational Efficiency and Operational Metrics of DocParseNet}
\label{appendix2-Computational}

Table \ref{table3-appendix} delineates the computational efficiency and operational metrics of DocParseNet in relation to established semantic segmentation models. The table quantifies the models in terms of model parameters, memory usage, TFLOPs for one training step, and execution time per epoch, underpinning the substantial computational advantages of DocParseNet. With only 2.8 million parameters, DocParseNet demonstrates an optimal balance between model complexity and performance latency, achieving a low average per-epoch execution time of 25 seconds. This is considerably faster than the other models, as DocParseNet uses 6.06 GB of CPU memory and 9.30 GB of GPU memory, which are within practical limits for deployment in real-world applications. These experiments, conducted with a batch size of 8, except for Segformer-B5, which uses a batch size of 4, showcase the model's ability to deliver high performance without the need for extensive computational resources.


\subsection{Examples of Annotated Scanned Documents}
\label{appendix2-sample}
\begin{figure}[H]
\renewcommand\thefigure{B3}
    \centering
    \subfloat[Annotated Sample 1]{%
        \includegraphics[width=0.65\textwidth,  trim=0 4cm 0 4cm]{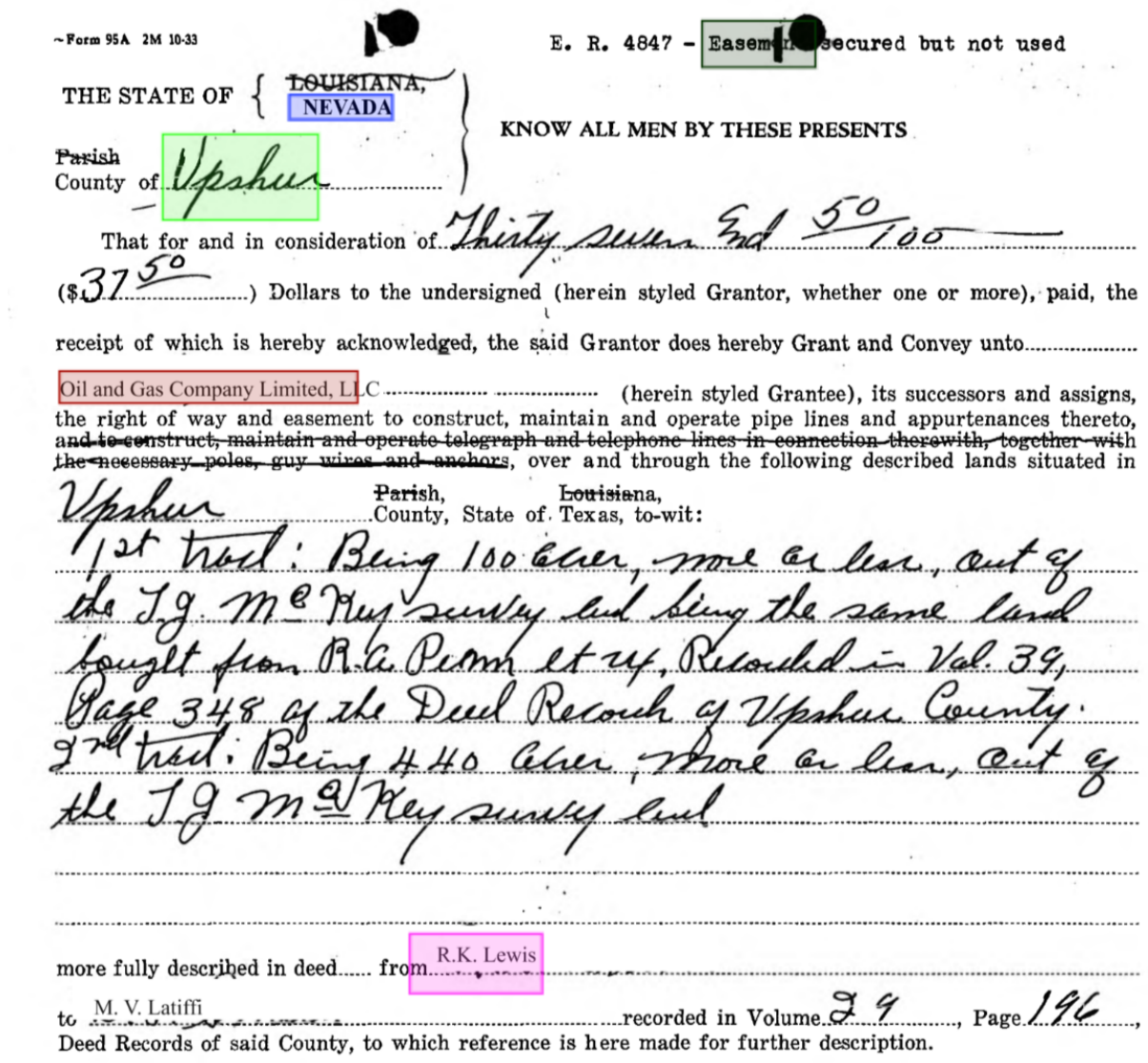}
        \label{fig:anno-1}
    }
    \vspace{-5mm} 
    \subfloat[Annotated Sample 2]{%
        \rotatebox{-90}{\includegraphics[width=0.58\textwidth]{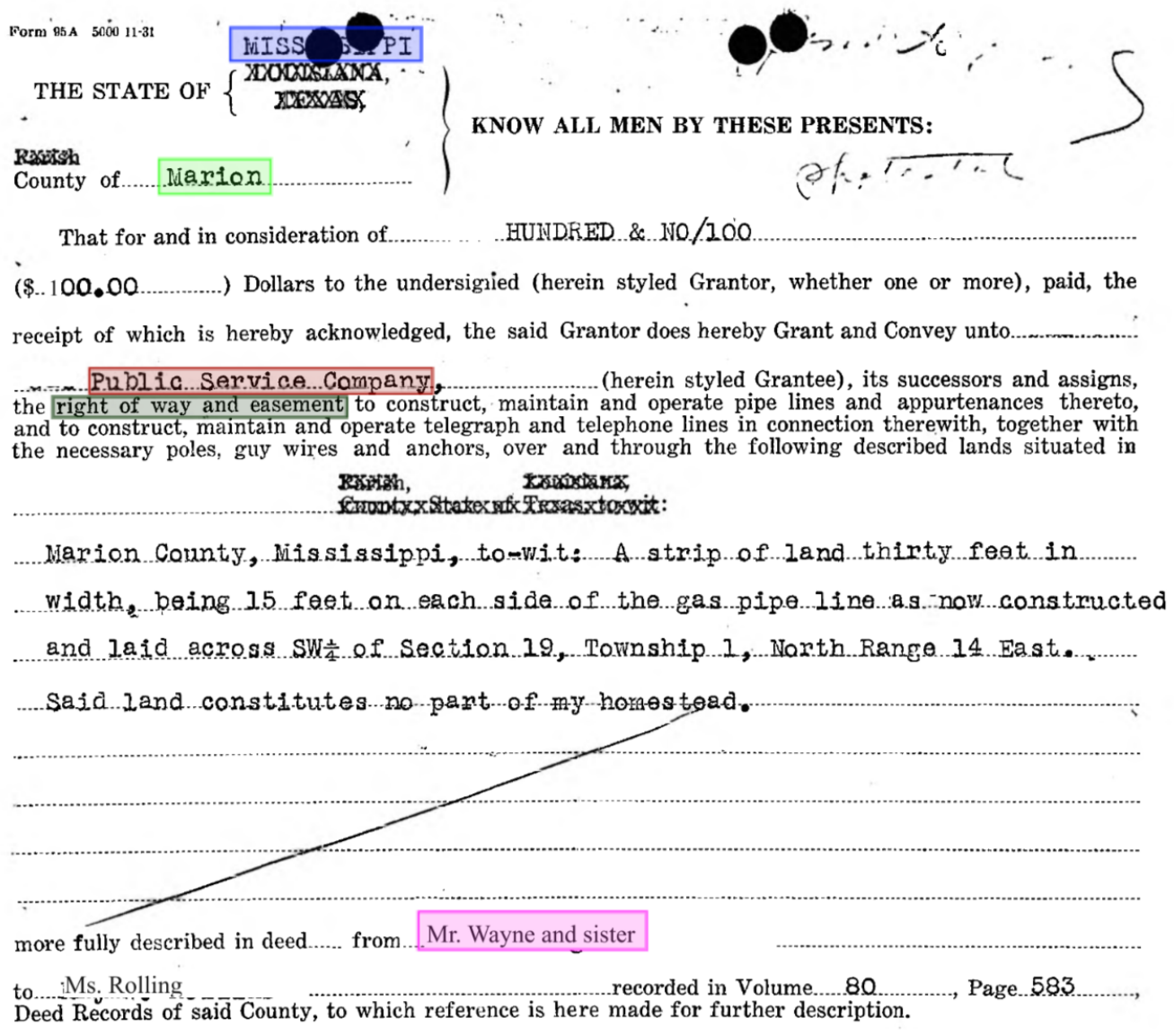}}
        \label{fig:anno-2}
    }
    \vspace{-2mm}
    \caption{Annotated samples demonstrating different perspectives}
    \label{fig:anno}
\end{figure}

\end{document}